\documentclass[runningheads]{llncs}
\usepackage{graphicx}
\usepackage{amsmath,amssymb} 
\usepackage{color}

\usepackage{array}

\usepackage[width=122mm,left=12mm,paperwidth=146mm,height=193mm,top=12mm,paperheight=217mm]{geometry}
\begin{document}
\pagestyle{headings}
\mainmatter

\title{Can Ground Truth Label Propagation from Video help Semantic Segmentation?} 

\titlerunning{Can Label Propagation help Semantic Segmentation?}

\authorrunning{Mustikovela, Yang, Rother}

\author{Siva Karthik Mustikovela, Michael Ying Yang, Carsten Rother}


\institute{Technische Universit\"at Dresden\\
	\email{ \{siva\_karthik.mustikovela,carsten.rother\}@tu-dresden.de}\\
	University of Twente\*\\
	\email{ \{michael.yang\}@utwente.nl}
}

\maketitle

\begin{abstract}
For state-of-the-art semantic segmentation task, training convolutional neural networks (CNNs) requires dense pixelwise ground truth (GT) labeling, which is expensive and involves extensive human effort.  In this work, we study the possibility of using auxiliary ground truth, so-called \textit{pseudo ground truth} (PGT) to improve the  performance. The PGT is obtained by propagating the labels of a GT frame to its subsequent frames in the video using a simple CRF-based, cue integration framework. Our main contribution is to demonstrate the use of noisy PGT along with GT to improve the performance of a CNN. We perform a systematic analysis to find the right kind of PGT that needs to be added along with the GT for training a CNN. In this regard, we explore three aspects of PGT which influence the learning of a CNN: i) the PGT labeling has to be of good quality; ii) the PGT images have to be different compared to the GT images; iii) the PGT has to be trusted differently than GT. We conclude that PGT which is diverse from GT images and has good quality of labeling can indeed help improve the performance of a CNN. Also, when PGT is multiple folds larger than GT, weighing down the trust on PGT helps in improving the accuracy. Finally, We show that using PGT along with GT,  
the performance of Fully Convolutional Network (FCN) on Camvid data is increased by $2.7\%$ on IoU accuracy.
 We believe such an approach can be used to train CNNs for semantic video  segmentation where sequentially labeled image frames are needed. To this end, we provide recommendations for using PGT strategically for semantic segmentation and hence bypass the need for extensive human efforts in labeling.
\end{abstract}

\section{Introduction}
\label{sec::Intro}

Semantic segmentation is an extensively studied problem which has been widely addressed using convolutional neural networks (CNNs) recently. CNNs have been shown to perform extremely well on datasets such as Pascal VOC \cite{Everingham:voc}, NYU-D \cite{Silberman:ECCV12}, CityScapes \cite{Cordts2016Cityscapes}, etc. For efficient performance of CNNs, there are certain characteristics of training data which are required: i) the ground truth (GT) training data needs dense pixelwise annotations which requires an enormous amount of human effort. For instance, an image in the Cityscapes dataset takes about $1.5h$ for dense annotation \cite{Cordts2016Cityscapes}, ii) the training data has to be diverse in the sense that highly similar images do not add much information to the network. Such diversity in training data helps better modelling of the distribution of test scenarios. 

For semantic video segmentation, continuous annotation of consecutive frames is helpful rather than annotations of discrete and temporally separated frames. In such a case it is again extremely expensive to obtain dense pixelwise annotation of consecutive images in the video. To this end, we arrive at an important question:  can auxiliary ground truth training data obtained by using label propagation help in better performance of a CNN-based semantic segmentation framework? 

In this work, we explore the possibility of using auxiliary GT, to produce more training data for CNN training. 
We use the CamVid dataset \cite{Brostow2008} as an example, which contains video sequences of outdoor driving scenarios. But the methodology can be easily applied to other relevant datasets. The CamVid has training images picked at $1fps$ from a $30fps$ video, leading to one GT training frame for every $30$ frames. We propagate the GT labels from these images to the subsequent images using a simple CRF-based, cue integration framework leading to \textit{pseudo ground truth (PGT)} training images. It can be expected that the new PGT is noisy and has lower quality compared to the actual GT labeling as a result of automaitc label propagation. 
We train the semantic segmentation network FCN \cite{long_shelhamer_fcn_2015} using this data. In this regard, we explore three factors of how the PGT has to be used to enhance the performance of a CNN. 
\begin{enumerate}
    \item \textit{Quality} - The PGT labeling has to be of good quality in the sense that there should not be too much of wrong labeling.
    \item \textit{Diversity} - The PGT training images have to be different compared to the GT images, in order to match the potential diverse test data distribution. 
    \item \textit{Trust} - During the error propagation, the PGT has to be weighted with a trust factor in the loss function while training.
\end{enumerate}

Further, we systematically analyze the aforementioned dimensions through extensive experimentation to find the most influential dimension which improves the performance of the CNN. We perform experiments with two main settings. First, where equal number of PGT and GT training samples are present. Second, the number of samples of PGT is multiple folds larger than GT training samples. Our baseline is obtained by training the FCN only on the GT training images which stands at $49.6\%$. From our experiments, we have found that adding PGT to the GT data and training the FCN helps in enhancing the accuracy by $2.7\%$ to $52.3\%$.

The main contributions of this work are:
\begin{itemize}
    \item We perform exhaustive analysis to find the influential factors among \textit{Quality, Diversity} and \textit{Trust} which affect the learning in the presence of PGT data. We conclude that PGT images have to be diverse from the GT images in addition to their labeling to be of good quality. Trust on PGT data should be sufficiently low when there is multiple folds of PGT than GT data.
    \item We provide application specific recommendations to use PGT data, taking the above factors into account. In the case of semantic video  segmentation, when PGT is multiple folds larger than GT, it is advisable to have a low trust on PGT data. In case of image semantic segmentation, diverse high quality PGT data helps in improving the performance. 
\end{itemize}
Detailed discussions are further presented in experiments section (sec. \ref{sec:exp}).



\section{Related Work}
\label{sec:related}


Semantic video segmentation has received growing interest in the last few years, as is witnessed by its increasing number of works both in  foreground/background segmentation 
\cite{NagarajaSB15,FragkiadakiCVPR12,Godec2011,Lee2011,Li2013,Ochs2011,Papazoglou2013,Tang2013,Zhang2013,jain2014supervoxel,Giordano2015,Wu2015,Yang2015}
and multi-class semantic segmentation
\cite{BadrinarayananGC10,BadrinarayananIJCV13,BadrinarayananBC13,Liu2015}.
We will focus our review on the later.


The influential label propagation method in \cite{BadrinarayananGC10} jointly models appearance and semantic labels using a coupled-HMM model  in video sequences. 
This method was extended to include correlations between non-successive frames using tree structured graphical models in \cite{BadrinarayananBC13}.
\cite{BadrinarayananIJCV13} presents a mixture of temporal trees model for video segmentation, where each component in the mixture connects superpixels from the start to the end of a video sequence in a tree structured manner.
While \cite{BadrinarayananBC13,BadrinarayananIJCV13} adopt semi-supervised learning fashion for learning the pixel unaries, our method is principally  different to their approach.
In \cite{BadrinarayananBC13,BadrinarayananIJCV13}, they first set the pixel unaries to uniform distributions, use inference technique to estimate the pixel marginal posteriors, and then do iterative inference. 
We first generate PGT to train a neural network using combined GT and PGT data and perform the forward pass for the inference.
Furthermore, since dynamic objects are ignored in the evaluation of \cite{BadrinarayananBC13,BadrinarayananIJCV13}, we don't know how good their approach applies to these object classes.
While it has been shown experimentally in \cite{BadrinarayananBC13,BadrinarayananIJCV13} that unaries learned in semi-supervised manner can help improve segmentation accuracy, 
we have performed thorough analysis of using PGT.
%
\cite{HartmannGHTKMVERS12} proposes to learn spatiotemporal object models, with minimal supervision, from large quantities of weakly and noisily tagged videos.
In \cite{KunduLDLR14}, the authors propose a higher order CRF model for joint inference of 3D structure and semantic labeling  in a 3D volumetric model.
\cite{Liu2015} proposes an object-aware dense CRF model for multiclass semantic video segmentation, which jointly infers supervoxel labels, object activation and their occlusion relationship. 
Unlike aforementioned methods, we use PGT data for learning the CNN model to perform the inference.

Recently, CNNs are driving advances in computer vision, such as image classification \cite{krizhevsky_cnn_2012}, detection \cite{zhang_rcnn_2014,gupta_ECCV14}, recognition \cite{agrawal_nn_2014,oquab_cnn_2014}, semantic segmentation \cite{girshick2014rcnn,long_shelhamer_fcn_2015}, pose estimation \cite{toshev_pose_2014}, and depth estimation \cite{EigenPF14}.
The success of CNNs is attributed to their ability to learn rich feature representations as opposed to hand-designed features used in previous methods. 
%
In \cite{MobahiCW09}, the authors propose to use CNN for object recognition exploiting the temporal coherence in video. 
Video acts as a pseudo-supervisory signal that improves the internal representation of images by preserving translations in consecutive frames.
%
\cite{NIPS2015_5849} proposes a semi-supervised CNN framework for text categorization that learns embeddings of text regions with unlabeled data and labeled data.
%
A number of recent approaches, including Recurrent CNN \cite{pinheiro2014}  and FCN \cite{long_shelhamer_fcn_2015}, have shown a significant boost in accuracy by adapting state-of-the-art CNN based image classifiers to the semantic segmentation problem.
In \cite{PapandreouCMY15} the authors propose to build their model on FCN and develop EM algorithms for semantic image segmentation model training under weakly supervised and semi-supervised settings. 
The authors show that their approach achieves good performance when combining a small number of pixel-level annotated images with a large number of image-level or bounding box annotated images.
%
This has been confirmed by 
\cite{PathakKD15,XiaoXYHW15}, where \cite{XiaoXYHW15} addresses the problem of training a CNN classifier with a massive amount of noisy labeled training data and a small amount of clean annotations for the image classification task, and \cite{PathakKD15} propose to use image-level tags to constrain the output semantic labeling of a CNN classifier in weakly supervised learning fashion.
For semantic video segmentation, since obtaining a massive amount of densely labeled GT data is very expensive and time consuming, we analyze the usefulness of auxiliary training data. 
%
In \cite{YaoTCBPLC15}, the authors propose an interesting approach that  takes into account both the local and global temporal structure of videos to produce descriptions, which incorporates a spatial temporal 3D CNN representation of the short temporal dynamics. 
One could extend current 2D CNN model to 3D temporal CNN for video segmentation, but this is out of the scope of this paper. 


\section{Our Approach}
\label{sec:approach}

In this section we discuss the details of our approach to generate PGT data, sorting schemes of PGT data and training the CNN.

\subsection{Pseudo Ground Truth Generation using Label Propagation}
\label{sec:pgt_gen}

\begin{figure}
\centering
\includegraphics[width=120mm]{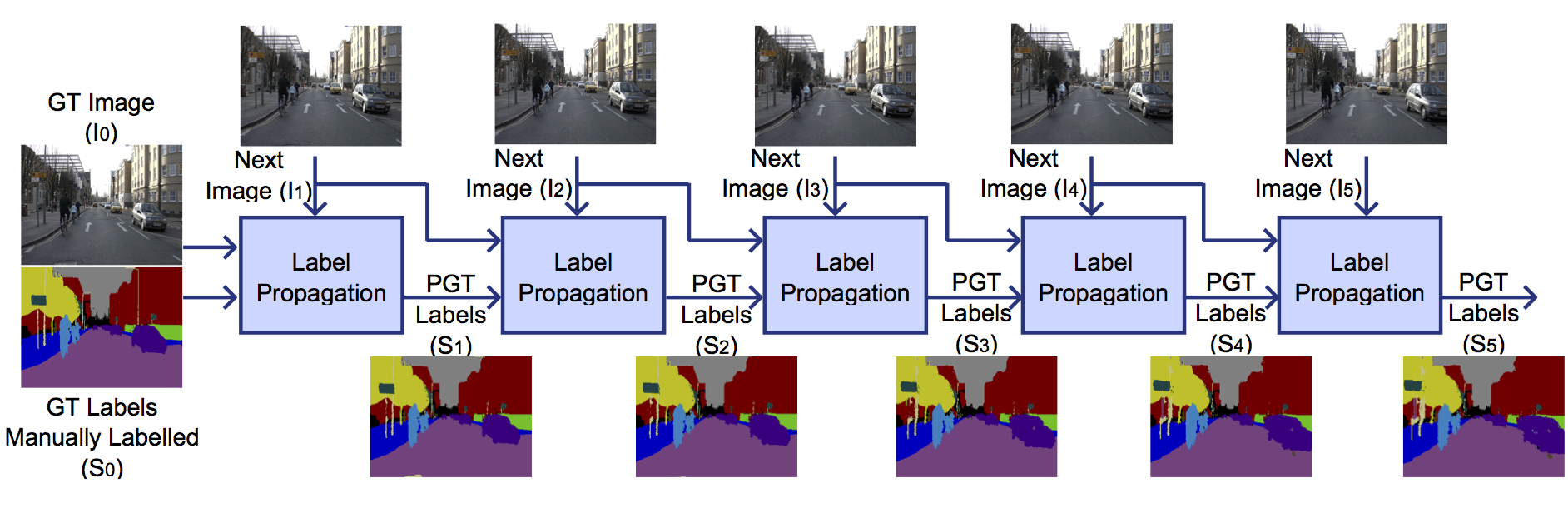} 
\caption{Illustration of generating Pseudo Ground Truth data.}
\label{fig:flow_chart}
\end{figure}

\paragraph{CamVid Dataset.}
The CamVid \cite{Brostow2008} dataset consists of video sequences of outdoor driving scenarios which is the most suitable dataset for our setting. It consists of 11 semantic classes namely \textit{Building, Tree, Sky, Car, Sign, Road, Pedestrian, Fence, Pole, Sidewalk, Bicycle}. The video sequences in this dataset are recorded at $30fps$. We use these sequences to extract individual image frames at the same rate for Pseudo Ground Truth Generation. The Ground Truth labeling exists for $1$ frame of every $30$ frames per second. This frame can be leveraged to propagate the GT labels of that frame to the following frame using the approach described below (Fig.~\ref{fig:flow_chart}). The training set for CamVid contains $M$ images ($M=367$).

\paragraph{Pseudo Ground Truth generation. }
Given a Ground Truth labeling $S\textsuperscript{t}$ for a frame $I\textsuperscript{t}$ in the training set, we propagate the semantic labels $S\textsuperscript{t+1}$ of that frame  to the next frame $I\textsuperscript{t+1}$ in the sequence. The labeling of this new subsequent frame in the sequence, $S\textsuperscript{t+1}$ is called \textit{Pseudo Ground Truth (PGT)}. We follow an approach similar to \cite{corso_label_prop}, but use additional smoothness terms and a different inference scheme. A graphical model is formulated using optical flow and texture cues. The labeling with optimal energy corresponds to the PGT $S\textsuperscript{t+1}$ of the considered subsequent frame $I\textsuperscript{t+1}$. The energy of the graphical model is defined as follows,
\begin{equation}
\label{eq::crf_energy}
E(S^{t+1} \mid S^t, I^t, I^{t+1}) =  
U^M(S^{t+1}, S^t, I^t, I^{t+1})  +  \lambda_1 U^C(S^{t+1},I^{t+1}) + \lambda_2 V^s (S^{t+1},I^{t+1})
\end{equation}
where $U$ and $V$ denote the unary and pairwise potentials in the model. The motion unary term $U^{M}$ defines the potential when a pixel $z{_{n}{^{t+1}}}$ in $I\textsuperscript{t+1}$ takes a label from one of the incoming flow pixels $z{_{n'}}{^{t}}$ from $I^t$ whose flow terminates at $z{_{n}{^{t+1}}}$. 
The motion unary is defined as
$$U^{M}(S^{t+1}, S^{t}, I^{t}, I^{t+1}) = \sum_{n}^{} \sum_{n' \mid z{_{n'}}{^{t}} \in f(z{_{n}{^{t+1}}})} w(z{_{n'}}{^{t}}, z{_{n}{^{t+1}}})(1-\delta(S{_{n}{^{t+1}}},S{_{n'}{^{t}}}))$$
where $\delta$ is the Kronecker delta and $f(z{_{n}{^{t+1}}})$ is the set of pixels in $I^t$ which have a forward optical flow terminating at $z{_{n}{^{t+1}}}$. The function $w$ defines the similarity of RGB histograms between two small image patches around $z{_{n}{^{t+1}}}$ and $z{_{n'}}{^{t}}$, measured using KL-Divergence.
The appearance unary term $U^C$ computes the log probability of a pixel $z{_{n}{^{t+1}}}$ belonging to one of the label classes. We learn a Gaussian Mixture Model for the texture of each semantic class using only the GT labeling of the first image($I^0$) of that particular sequence. 
$U^C$ is defined as
$$U^{C}(S^{t+1} \mid I^0, S^0) = \sum_{n} -\log P(z{_{n}{^{t+1}}} \mid \mu_0, \textstyle{\sum_0}) $$
where $\mu_0$ is the mean and $\sum_0$ is the variance of the GMM  over $I_0$. $P$ gives the likelihood of $z{_{n}{^{t+1}}}$ belonging to a certain class.   The pairwise term $V^s$ is a generic contrast sensitive Potts model. $V^s$ is given by
$$V^s(S^{t+1}, I^{t+1}) = \sum_{z_m,z_n\in c} dis(m,n)^{-1}[s_{z_m} \neq s_{z_n}] \exp(-\beta (h_{z_m}-h_{z_n})^2)$$
where $z_m$, $z_n$ are two connected pixels, $dis(.)$ gives the euclidean distance between $m$ and $n$, $[\phi]$ is an indicator function which takes values $0$, $1$ depending on the predicate $\phi$, $s_{z_m}$ is the label of $z_m$ and $h_{z_m}$ is the color vector of $z_m$. $\beta$ is a constant chosen empirically.  The inference is performed using mean-field approximation \cite{KrahenbuhlK11}.

We start the label propagation using the reference frame as $I^0$, which is the Ground Truth training frame, and $S^0$, its corresponding semantic labeling. The labels are propagated till the $5^{th}$ frame in the sequence. Figure~\ref{fig:flow_chart} gives an illustration of the label propagation. The labeling $S^{t+1}$ obtained after the inference is the so-called \textit{Pseudo Ground Truth} labeling for the frame $I^{t+1}$ using the reference labeling for $I^t$. In this way, for every GT labeled frame in the $367$ image training set, we propagate the labeling to the next $5$ consecutive frames obtained from the sequence. To this end, the total number of PGT frame labelings obtained is $1835$. We can as well propagate the labels to the backward frames. We assume that we get similar kind of PGT and experimental results in such a case as well.

\begin{figure}
\centering
\includegraphics[width=120mm]{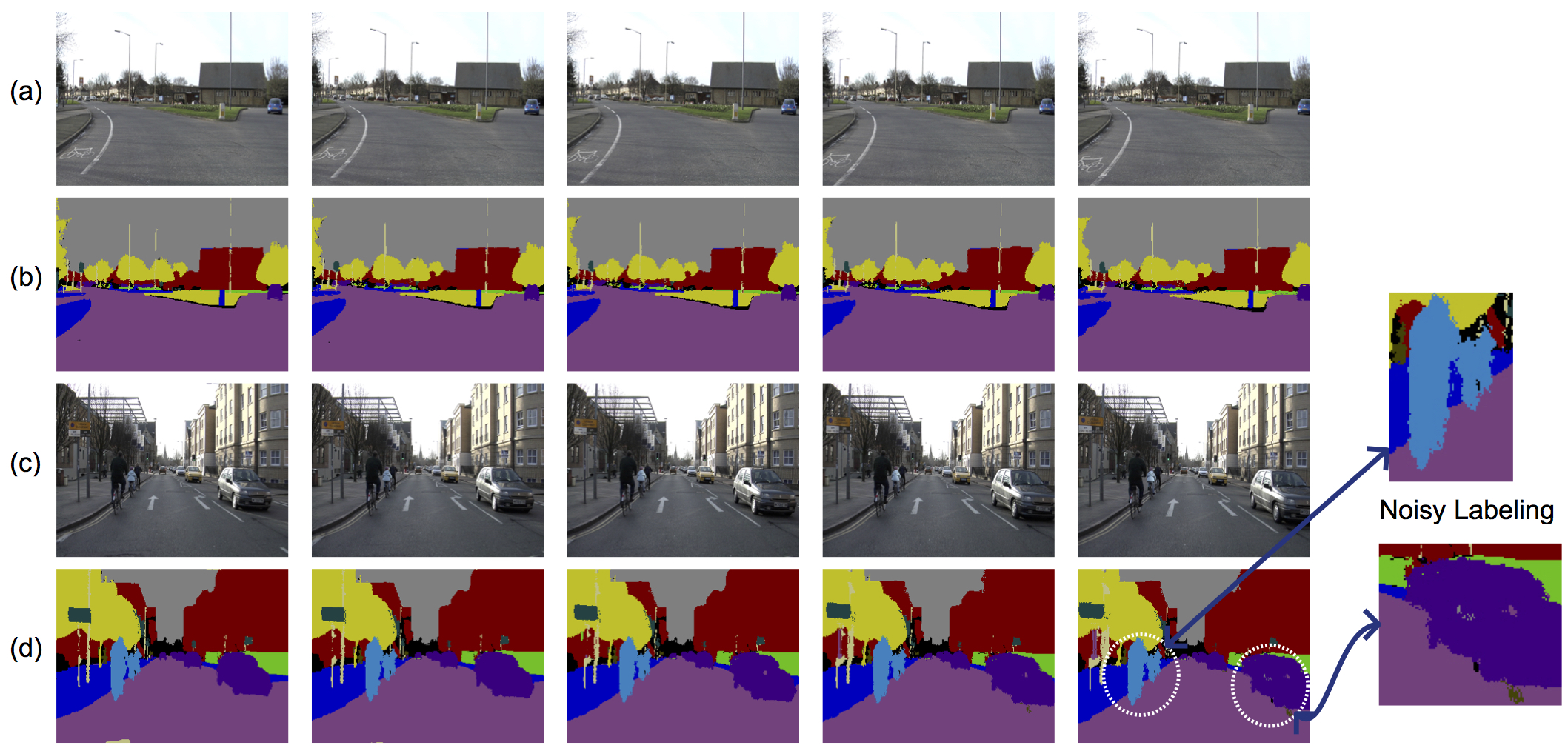} 
\caption{Quality of some PGT labelings. (a), (c) are sequence of images. (b), (d) are corresponding labelings. (b) shows an example of good quality PGT label propagation for a sequence, while (d) shows a case where there is much noisy labeling in the PGT. A zoomed-in version of the last image in the sequence (d) illustrates the noisy labeling.}
\label{fig:flow_chart_2}
\vspace{-0.7cm}
\end{figure}

\vspace{-0.3cm}

\subsection{PGT Data}
\label{sec:pgt_data}
The focus of this work is to determine the right kind of PGT data that needs to be added to the GT to enhance the performance of CNN. As mentioned before, there are three factors of considering a PGT labeling while learning the CNN.
\begin{enumerate}
    \item The quality of the labeling should be good. As it can be seen from Fig.~\ref{fig:flow_chart_2}(d), the PGT labeling for an image can be erroneous. On the other hand, labeling can also be reliable enough (Fig.~\ref{fig:flow_chart_2}(b)). This presents a situation where the right kind of high quality PGT must be chosen. 
    \item An important requirement in the learning of CNNs is that the data should be as diverse as possible. This aids in better modeling of the distribution of test images, which we assume are diverse. Hence, the images appearing in the later part of the sequence can be expected to enhance the performance of the network. 
    \item Because of the noise in the PGT labeling, the data may not be completely trusted when training a CNN for semantic segmentation. Hence, the gradient obtained during back propagation for a PGT has to be scaled appropriately.
\end{enumerate}
The above mentioned criteria form the basis for sorting the PGT data using various schemes as mentioned below.

\begin{figure}[h]
\centering
\includegraphics[width=60mm]{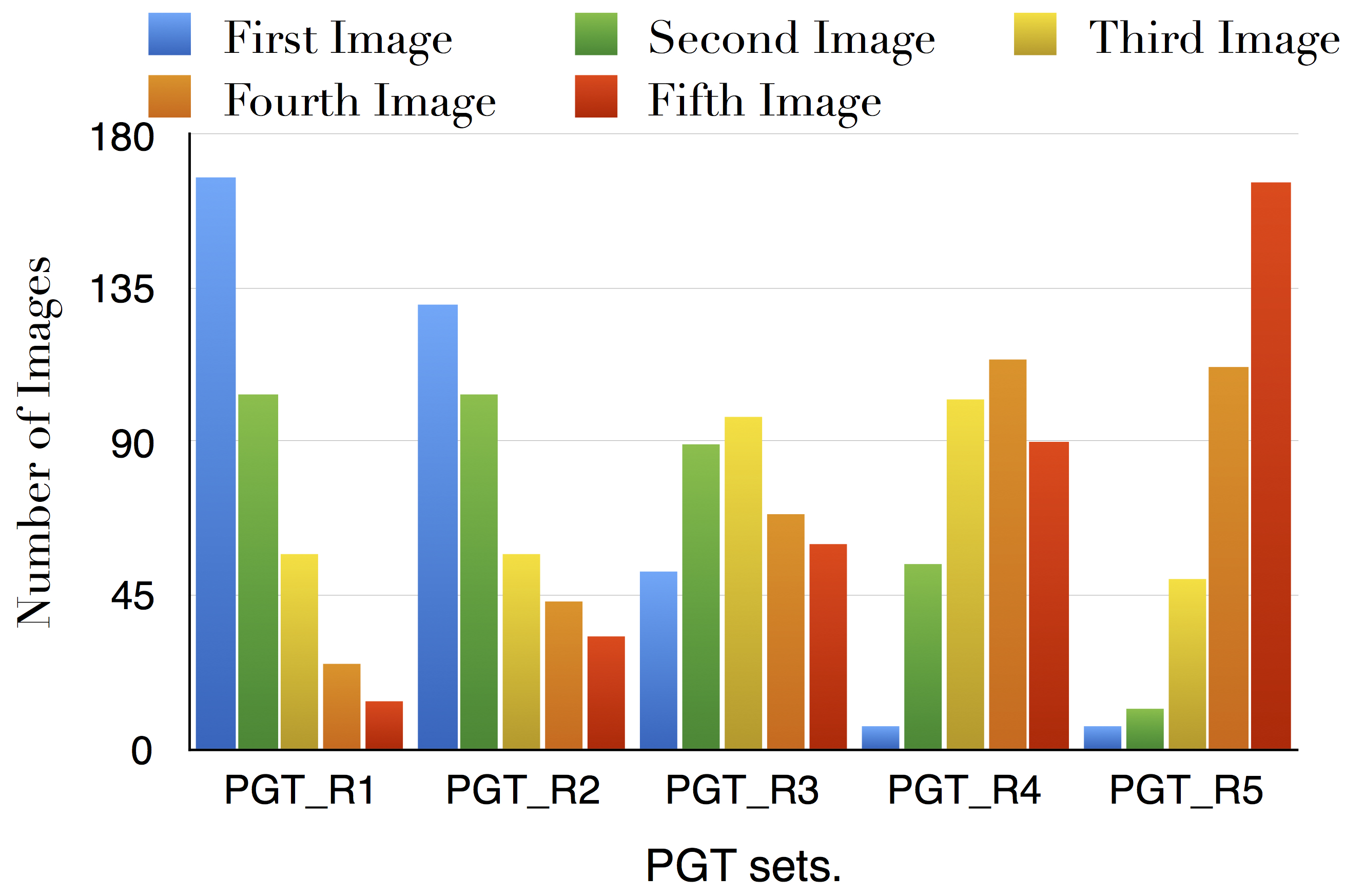} 
\caption{Distribution of sequence of images in each visually sorted set.}
\label{fig:ranked_dist}
\end{figure}

\paragraph{Visual rating based on labeling quality.}
The quality and the noise in the PGT labeling differs from image to image (Fig.~\ref{fig:flow_chart_2}) and we would like to choose the PGT data with high label quality. To achieve this, the PGT labelings are manually rated, based on their visual quality, ranging from $1$ to $9$. The labelings are checked for class label consistency, e.g label drift from one semantic region to another. We observed that the labeling quality goes down as we move away from the GT labeling.  
All the GT image labelings are rated as $10$ which serve as the baseline. Further, all the PGT labelings are sorted according to their rating and distributed into $5$ different sets (PGT\_R1 to PGT\_R5) each containing $367$ labelings. For instance the first set, PGT\_R1 contains the highest quality $367$ labelings and PGT\_R2 contains the next best high quality $367$ labelings. It could be expected that the first set PGT\_R1 majorly contains the first images in the sequence since they are generally of higher quality. Figure~\ref{fig:ranked_dist} shows the distribution of chronological images in each set. We call these as visually sorted sets for future reference. Rating $1835$ images took about about 2h ($4$-$5$ sec/image) for a human. Dense labelling of $1835$ images would take about ~1000h which is extremely expensive. We believe that the performance enhancement achieved at the expense of minimal human effort here is valuable.

\paragraph{Sequential Grouping.}
In this scheme, we group all the PGT labelings by their chronological order (PGT\_S1-PGT\_S5). For instance, PGT\_S1 contains all the $1^{st}$ labelings in the sequence and PGT\_S2 contains all the $2^{nd}$ labelings in the sequence. We call these as sequential sets. In this case PGT\_S4, PGT\_S5 can be expected to contain the most diverse images compared to the GT images because they appear later in the sequence.

\subsection{Trust on PGT while learning}
\label{sec::pgt_trust}
As mentioned in Sec.~\ref{sec:pgt_data}, the PGT data cannot be completely trusted when it is being learnt. To address this, we scale the gradient obtained while back propagating the error through the CNN by a trust factor less than $1$.  Let the trust factor for a PGT labeling be $t_f$. The new update rule for the Stochastic Gradient Descent is then
\begin{eqnarray}
\theta' := \theta - \eta t_{f} \triangledown J(\theta) \quad if \quad S \in PGT \\
\theta' := \theta - \eta \triangledown J(\theta) \quad if \quad S \in GT
\end{eqnarray}
where $\theta$ are the weights of the network, $\eta$ is the learning rate and $\triangledown J(\theta)$ is derivative of the loss function and $S$ is the semantic labeling upon which the loss is being computed. Effectively, the trust factor scales the magnitude of the gradient because the direction of the gradient cannot be completely relied upon, when learning a PGT image. In the experiments, we used trust factors varying from $0.5$ to $1$. A value of $t_f=1$ means that the PGT is trusted completely. All the GT labelings have a trust factor of $1$ by default. We also tried using a separate trust factor for each class by scaling the trust factor by the number of pixels of a certain object label. Since, it did not show any improvement in our experiments, we decided to use a single constant.

\subsubsection{Training}
We use the standard implementation of FCN \cite{long_shelhamer_fcn_2015} in all the experiments. We train the FCN using SGD with momentum and our modified update rule. We use a batch size of 1, with learning rate of $1^{-9}$ with a momentum 0.99 and weight decay of $5^{-4}$. We initialize the FCN with weights of VGG-16 network.
For the whole set of experiments, we need 60 GPUs for about 210h. Requirement of such resources is a major hindrance and hence we limit to experiments using only 5 consecutive frames.

\section{Experiments}
\label{sec:exp}
\textbf{}
In this section, we evaluate our approach using various experiments and further present a discussion of the results. As explained above, we consider two schemes for sorting PGT data, i.e. through visual sorting and sequential sorting. Below we discuss two different experimental setups, when adding the 5 sets individually to the GT, and when accumulating the 5 sets with the GT.

\subsection{Overall Performance}
We first train the FCN only using the GT Training data, consisting of 367 images. This gives an average class IoU accuracy of $49.6\%$. We consider this as the reference baseline. Table \ref{tab:comparison} gives a detailed comparison of all the class IoU accuracies and the average IoU.
From Table \ref{tab:comparison}, it can be seen that the FCN trained with extra PGT outperforms the FCN trained only on GT in all the classes. In this case, the PGT set is PGT\_S4 with trust factor = $0.9$. The average IoU accuracy is $52.3\%$ which is higher compared to $49.6\%$ in case of baseline FCN. For classes like \textit{Bicycle, Sidewalk, Fence, Car} our model provides a commendable performance boost. 
 Figure~\ref{fig:qualitative_results} shows some qualitative results of labeling obtained on test frames by our trained model.

\footnotesize
\begin{table}[]
\centering
\caption{Semantic segmentation performance of our approach compared to an FCN trained only on GT data.}
\label{tab:comparison}
\begin{tabular}{|c|c|c|c|c|c|c|c|c|c|c|c|c|}
\hline
Approach                                                   & \begin{tabular}[c]{@{}c@{}}Build\\ ing\end{tabular} & Tree & Sky  & Car           & Sign & Road          & \begin{tabular}[c]{@{}c@{}}Pedes\\ trian\end{tabular} & Fence         & Pole & \begin{tabular}[c]{@{}c@{}}Side\\ walk\end{tabular} & \begin{tabular}[c]{@{}c@{}}Bicy\\ cle\end{tabular} & \begin{tabular}[c]{@{}c@{}}Avg \\ IoU\end{tabular}                                              \\ \hline
\begin{tabular}[c]{@{}c@{}}FCN\\ (Only GT)\end{tabular}   & 70.5                                                & 63.1 & 84.8 & 61.9          & 19.1 & 89.8          & 19.8                                                  & 30.9          & 6.5  & 70.1                                                & 29.3                                               & 49.6                                               \\ \hline
\begin{tabular}[c]{@{}c@{}}Ours\\ (GT+PGT\_S4 \\ $t_f$=0.9)\end{tabular}  & \textbf{72}                                         & \textbf{65.6} & 84.6 & \textbf{64.6} & \textbf{20.8} & \textbf{90.6} & \textbf{24.9}                                                  & \textbf{38.8} & \textbf{8.0}  & \textbf{71.8}                                       & \textbf{33.9}                                      & \textbf{52.3}                                      \\ \hline
\end{tabular}
\vspace{-0.4cm}
\end{table}

\subsection{Effect of ambiguous labeling of images}
\label{sec:amb_lab}
In this experiment we want to analyze the effect of ambiguity in the semantic labeling of extremely similar images. We performed a preliminary experiment in which the PGT is produced just by using the labels of the first frame directly for all the successive frames. This is the most naive way of propagating labels from the base frame to the next 5 frames. We trained the FCN with GT and PGT containing either all the first frames or all second frames etc. We observed the the accuracy sharply decreases as we move away from GT image in the following manner - 50.7 (GT+1st frames), 50.6 (GT+2nd frames), 50.2, 49.9, 49.7 (GT+ 5th frames). This is because the labeling that comes from the first frame GT is more erroneous when applied to the later frames. 

Further, we performed another experiment where the labels of the first image are jittered and applied to the same image again. Unlike the above experiment, jittered labels are applied to the same image. To achieve this, we dilate the object labels along their borders by a very small amount followed by a minor random shift of 2-4 pixels. This simple technique creates an ambiguous labeling along the borders, effectively leading to a jitter of all semantic object labels. Effectively, such ambiguous labels mimic the labels that are a result of label propagation where they are generally erroneous, at the borders of the semantic objects. For each image-labels pair $I_n$-$S_n$ in the GT set, we create 3 such ambiguous labelings ($I_{n}$-$S_{n}^1$, $I_{n}$-$S_{n}^2$, $I_{n}$-$S_{n}^3$). Further, three training sets are created (AGT\_1, AGT\_1-2, AGT\_1-3).  AGT\_1 contains the GT training data and the set of all first ambiguous labelings for each image. Likewise, AGT\_1-3 contains the GT data, and all the ambiguous labelings for each image. Essentially, we are trying to analyse the effect of ambiguous labeling for a set of extremely similar images. 
We train the FCN using various trust factors for PGT data ranging from $0.5$ to $1$. From this experiment (last column of Table~\ref{jitter_table}), it can be observed that the addition of multiple folds of ambiguous labels for a set of extremely similar images reduces the accuracy (AGT\_1 to AGT\_1-3). This is because of the addition of extremely similar images to the CNN with more ambiguous labeling which corrupts the learning of CNN. This effect would further be useful in explaining certain phenomena in the experiments below. The average accuracies are marginally more than 49.6\% (baseline) for AGT\_1, AGT\_1-2. The reason is that, effectively, we are increasing the number of epochs for which the CNN is trained when we are using the same image set repeatedly.

\footnotesize
\begin{table}[]
\centering
\caption{Accuracies for various FCNs trained on ambiguously labeled data.}
\label{jitter_table}

\begin{tabular}{|c|c|c|c|c|c|c|c|}
\hline
\begin{tabular}[c]{@{}c@{}}Trust Factor/\\ Tr. Set\end{tabular} & 0.5  & 0.6  & 0.7  & 0.8  & 0.9  & 1    & \begin{tabular}[c]{@{}c@{}}Avg. \\ Acc.\end{tabular} \\ \hline
AGT\_1                                                          & 50.1 & 50.1 & 50   & 50.2 & 50.3 & 49.2 & 50.0                                                   \\ \hline
AGT\_1-2                                                        & 50.1 & 50   & 50.3 & 50.1 & 50.1 & 49.9 & 50.0                                                   \\ \hline
AGT\_1-3                                                        & 48.5 & 48.7 & 49.2 & 49.1 & 49.4 & 49.2 & 49.0                                                   \\ \hline
\end{tabular}
\vspace{-0.5cm}
\end{table}

\subsection{FCN trained using separate PGT sets}
\label{sec:FCN_sep_sets}

\subsubsection{(A) Sequentially sorted PGT sets.}
In this experiment, we train the FCN with training sets containing 734 images which consist of GT (367 images) and one of the PGT\_S training sets (367 images). As described in Sec.~\ref{sec:pgt_data}, there are $5$ PGT sets in this experiment, each consisting of $1$\textsuperscript{st}, $2$\textsuperscript{nd}, $3$\textsuperscript{rd}, $4$\textsuperscript{th} and $5$\textsuperscript{th} images in the sequence starting from the GT image. We train the FCN using various trust factors for PGT data ranging from $0.5$ to $1$. In all these experiments, the trust on the GT training set is fixed to be $1$. Table \ref{tab::sep_sets}(a) outlines the accuracies for each training set over various trust factors. Following are the observations of this experiment:

{\textbf{Trust Factors:}} From Table~\ref{tab::sep_sets}(a) it can be observed that there is no clear trend in the effect of trust factors on the learning of CNN, particularly when a single set of PGT is added to the training. For example, in the case of GT$+$PGT$1$, $t_f=0.6$ performs the best with an accuracy of $51.5$, while $t_f=0.5$ performs the best in case of GT$+$PGT$2$. In case of GT+PGT4, $t_f=0.9$ gives the highest accuracy of $52.3\%$ and the other trust factors lead to similar accuracies in the range of $51.3\%$ to $51.5\%$. This can be attributed to the extreme non-convexity of the function space of CNNs, due to which the scaling of magnitude of the gradient could lead to a different local minima. For this reason, to compare the effect of various training sets, we average the accuracies due to various trust factors (see last column in each table). 

\begin{figure}[h]
\centering
\includegraphics[width=95mm]{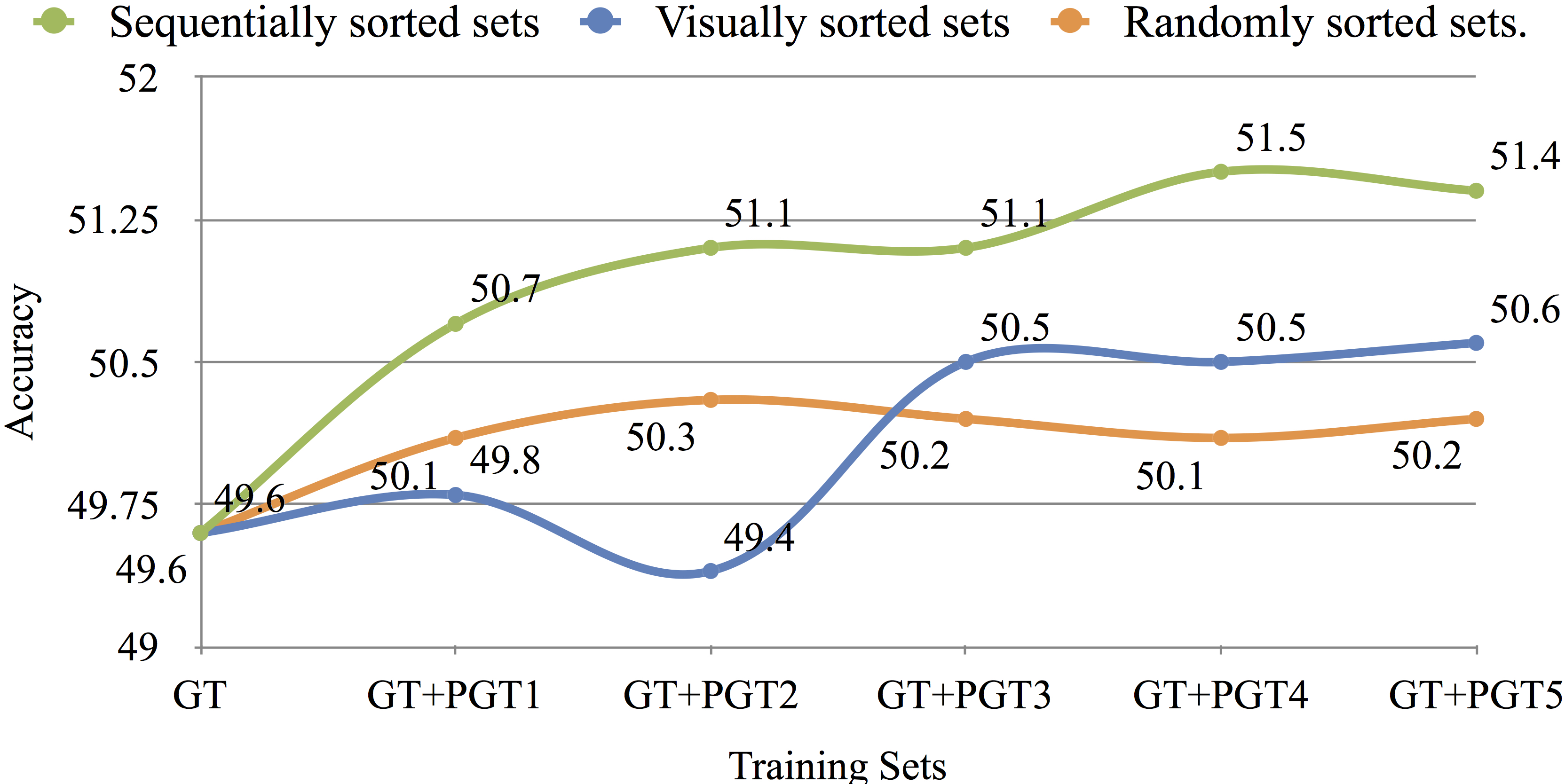} 
\caption{Graph shows the average accuracy change upon varying the PGT set added to training along with GT. Green curve indicates the accuracies in case of sequentially sorted sets. Blue curve indicates the accuracies in case of visually sorted sets. Orange curve indicates the accuracies for randomly sorted sets.}
\label{fig:sep_sets}
\vspace{-0.4cm}
\end{figure}

\begin{figure}
\centering
\includegraphics[width=120mm]{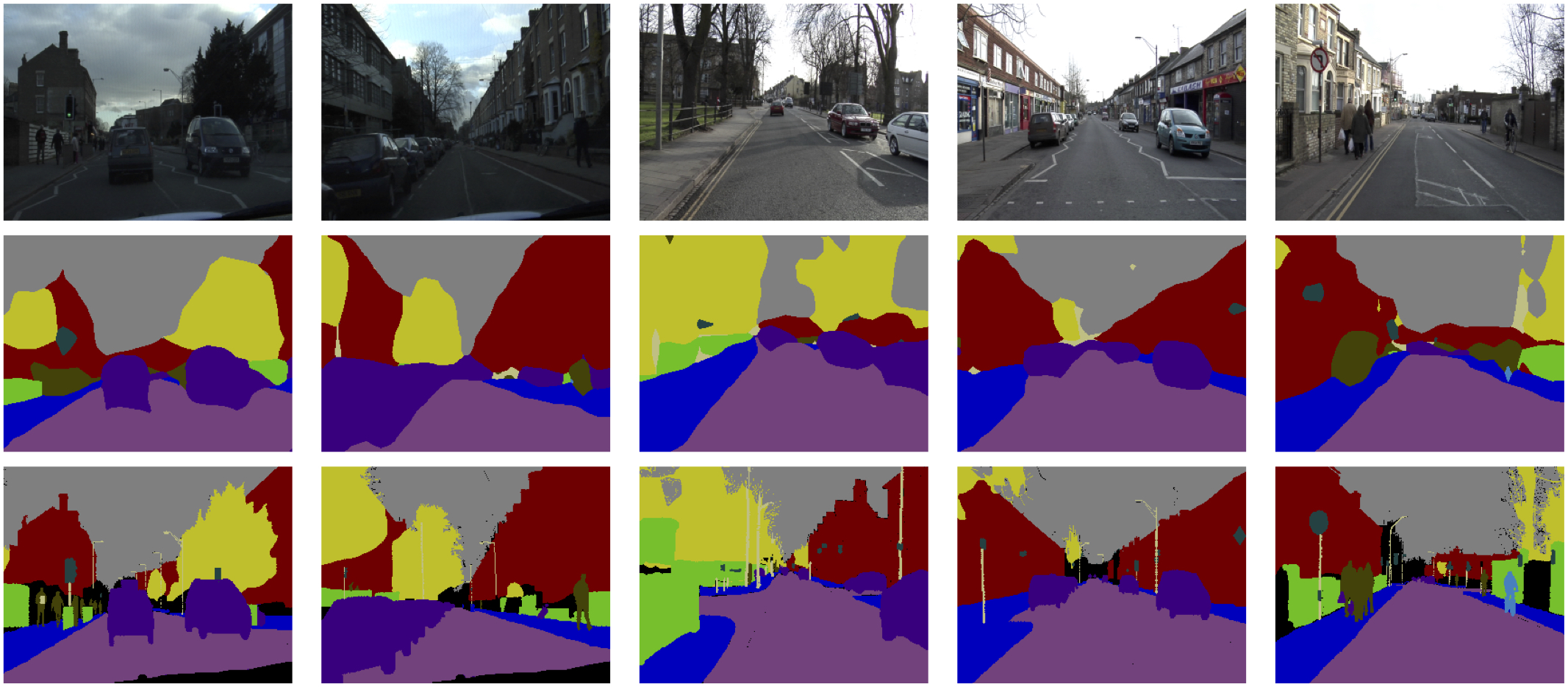} 
\caption{Qualitative performance of our system. First row-Images. Second row-Output of FCN trained with GT+PGT\_S4, trust factor = 0.9. Third row-ground truth.}
\label{fig:qualitative_results}
\end{figure}

{\textbf{PGT Sets:}} It can be seen that the accuracy for a training set averaged over trust factors is the highest ($51.5\%$) when the training set contains all the $4$\textsuperscript{th} images in the sequence. Further, the average accuracy declines to $51.4\%$ when the set of all the $5$\textsuperscript{th} images is included in the training set. Fig. \ref{fig:sep_sets} shows the trend of average accuracy of the network when different sequentially sorted PGT sets are added to the training data. It can be clearly seen that the accuracy increases as the PGT images are farther away from the GT images and further drops when set of $5^{th}$ images is added. This can be attributed to the fact that the PGT sets with $4^{th}$ and $5^{th}$ images are very different compared to those in the GT images and the labeling in the $5^{th}$ set is lower in quality. Effectively, the $4^{th}$ set aids in providing diverse high quality data and adding more information to the network. Effectively, if there is a better label propagation algorithm which can reliably propagate labels to farther images in the sequence, we conjecture that a later image would help in providing more information to the CNN and hence further enhance the performance of CNN.

From the above observations, it can be concluded that, in learning a CNN, the effect of using a high quality diverse PGT set (e.g. $4^{th}$ set) is more prominent than the effect of a trust factor on a PGT set, in a case where the number of PGT images is the same as the number of GT images in the training data.

\begin{table}[]

\centering
\caption{Accuracies for various trust factors of training the FCN with GT images and an additional set of equal number of PGT images.}
\label{tab::sep_sets}
\begin{tabular}{c|c|c|c|c|c|c|c|c|c|c|c|c|c|c|c|}
\multicolumn{1}{l}{}                                                           & \multicolumn{7}{c}{\textbf{(a)Accuracies for Sequential Sets}}                                                   & \multicolumn{1}{l}{} & \multicolumn{7}{c}{\textbf{(b)Accuracies for Rated Sets}}                                              \\ \cline{1-8} \cline{10-16} 
\multicolumn{1}{|c|}{\begin{tabular}[c]{@{}c@{}}Trust Fac./\\ Tr. Set\end{tabular}} & 0.5  & 0.6  & 0.7  & 0.8  & 0.9           & 1    & \begin{tabular}[c]{@{}c@{}}Avg.\\ Acc\end{tabular} &                       & 0.5  & 0.6  & 0.7  & 0.8  & 0.9  & 1    & \begin{tabular}[c]{@{}c@{}}Avg.\\ Acc\end{tabular} \\ \cline{1-8} \cline{10-16} 
\multicolumn{1}{|c|}{GT+PGT1}                                                   & 50.7 & 51.5 & 51.2 & 50.8 & 50.6          & 49.5 & 50.7                                               &                       & 49.7 & 49.3 & 48   & 50.2 & 50.9 & 50.7 & 49.8                                               \\ \cline{1-8} \cline{10-16} 
\multicolumn{1}{|c|}{GT+PGT2}                                                   & 52.1 & 51.2 & 50.2 & 51.5 & 50.4          & 51.3 & 51.1                                               &                       & 49.8 & 48.9 & 49.4 & 49.5 & 49.8 & 48.7 & 49.4                                               \\ \cline{1-8} \cline{10-16} 
\multicolumn{1}{|c|}{GT+PGT3}                                                   & 51.5 & 51.4 & 50.5 & 51.1 & 50.9          & 51.1 & 51.1                                               &                       & 51.2 & 50.7 & 51.2 & 50.5 & 50.3 & 49.4 & 50.5                                               \\ \cline{1-8} \cline{10-16} 
\multicolumn{1}{|c|}{GT+PGT4}                                                   & 51.4 & 51.4 & 51.3 & 51.3 & \textbf{52.3} & 51.3 & 51.5                                               &                       & 51.5 & 50.8 & 50.2 & 50.3 & 50.1 & 50.5 & 50.5                                               \\ \cline{1-8} \cline{10-16} 
\multicolumn{1}{|c|}{GT+PGT5}                                                   & 51.2 & 52.1 & 51.2 & 51.5 & 51.1          & 51.4 & 51.4                                               &                       & 50.9 & 50.9 & 50.1 & 50.1 & 51.5 & 50.6 & 50.6                                               \\ \cline{1-8} \cline{10-16} 
\end{tabular}
\vspace{-0.5cm}
\end{table}

\subsubsection{(B) Visually Rated PGT Sets.}
In this experiment, we train the FCN with training sets containing 734 images which consist of GT (367 images) and a PGT\_R training sets (367 images). PGT\_R1 contains the highest quality PGT labeling followed by PGT\_R2 which contains the next highest quality PGT labeling, and so on. We train the FCN using various trust factors for PGT data ranging from $0.5$ to $1$. The trust on the GT training set is fixed to be $1$. Table~\ref{tab::sep_sets}(b) outlines the accuracies for each training set over various trust factors. The following are the observations of this experiment. It can be seen that all 30 FCN models trained on GT+PGT perform better than our baseline which indicates that there is an improvement in performance of FCN when PGT data is used.

{\textbf{Trust Factors:}} There is no clear trend in the effect of trust factors on the learning of CNN. As mentioned above, the extreme non-convexity of the function space of CNN may prevent us from explicitly analyzing the effect of trust factors.  

{\textbf{PGT sets:}} The accuracy for a training sets averaged over trust factors increases as we present the PGT sets PGT\_R1 to PGT\_R5. The PGT labeling in PGT\_R5 is lower in quality compared to those in PGT\_R1, but the accuracy is higher for PGT\_R5 compared to that of PGT\_R1. This could seem to be counter intuitive, but upon further analysis the trend is explicable. Consider Fig.~\ref{fig:ranked_dist} where the constitution of chronological images in visually sorted sets is depicted. It can be observed that the sets starting from the PGT\_R3 majorly consist of later images than the earlier images in the sequence. Hence, the diversity of training data increases as we proceed to later sets. 

{\textbf{Label ambiguity:}} It can be seen that the average accuracies for visually sorted sets (Fig.~\ref{fig:sep_sets}) are lower than those of sequentially sorted sets. This might as well seem to be unreasonable because the PGT labeling in visually sorted sets are of high quality. However, there is an important detail to notice here. These initial images are extremely correlated and similar. The presence of multiple highly correlated images and label ambiguity for these brings down the accuracy. While in the case of sequentially sorted images, only one of the highly correlated images in the sequence is present for each image in the GT.  

From the above observations, it can be clearly concluded that the presence of diverse PGT images plays a stronger role in enhancing the accuracy of CNNs, rather than high quality images, particularly when the label propagation is done for shorter sequences. To this end, we conjecture that the quality of images could play a considerable role when the label propagation is done for longer sequences. Such an effect can already be seen in the case of sequentially sorted PGT sets where the accuracy for PGT\_S5 is less than that of PGT\_S4. This can be clearly attributed to the fact that the label quality in PGT\_S5 is less than that of PGT\_S4.

\subsubsection{(C) Randomly selected sets.}
In this setting, we divide the set of all the PGT data randomly into $5$ sets each containing $367$ images. An image-label sample belongs to only one set in these five sets. As it can be seen in Fig. 5(orange plot), the average accuracies do not majorly change when the PGT sets are changed. While in cases of sets $3,4,5$ of the sequential and visually sorted images, the accuracies are higher than those of randomly selected sets. This again is for the reason that there is no major steer of selection of diverse high quality labeling in a randomly assigned set. Clearly, this experiment again reinforces the necessity of selection of diverse good quality labeling to improve the performance of CNN.

\subsection{FCN trained using accumulated PGT sets}

In this section we describe the experiments where the PGT sets are accumulated for training the FCN (Table \ref{tab:acc_sets}). For instance, Set $1$ contains the GT, first PGT set and the second PGT set. The last set contains GT and all the PGT sets. Note that the ratio of PGT samples to the GT samples goes up as we proceed. So the last set contains $5$ times as many PGT samples as the GT samples. Similar to above section, we present two experiments where we accumulate sequentially sorted sets (Table~\ref{tab:acc_sets}(a)) and visually rated sets (Table \ref{tab:acc_sets}(b)). We train the FCN using various trust factors for PGT data ranging from $0.5$ to $1$. The trust on the GT training set is fixed to be $1$.

{\textbf{Trust Factors:}}
The last row in Table~\ref{tab:acc_sets} contains the average accuracy over all the datasets for a given trust factor. Unlike the earlier observation in Sec.~\ref{sec:FCN_sep_sets} where the trust factors analysis did not show a clear trend, here it can be seen that lower trust factors help to produce better accuracies compared to higher trust factors. As seen from the last row (Table~\ref{tab:acc_sets}(a)), the average accuracy constantly declines from $50.2\%$ to $49.2\%$ when the trust factor is varied from $0.5$ to $1$ for sequentially sorted sets. Also, it can be seen from the last row (Table~\ref{tab:acc_sets}(b)), that the average accuracy constantly declines from $50.6\%$ to $48.6\%$ when the trust factor is varied from $0.5$ to $1$ for visually ranked sets. Note that even in the presence of multiple folds of noisy PGT samples in the training set, the FCN performs at an average accuracy of $50.6\%$ when the trust factor is set to a sufficiently low value.

From the above observations, we conclude that the trust factor for the PGT data should be low enough when the number of PGT samples is considerably higher than that of GT samples.

\begin{table}[]
\centering
\caption{Accuracies of various training experiments when the PGT data is accumulated for each training. GT+(PGT1-5) contains all the PGT sets 1,2,3,4,5.}
\label{tab:acc_sets}
\begin{tabular}{|c|c|c|c|c|c|c|c|c|c|c|c|c|c|}
                                                                            \multicolumn{6}{c}{\textbf{{\begin{tabular}[c]{@{}c@{}}(a) Accuracies for \\ accumulated Sequential sets\end{tabular}}}} &  & \multicolumn{6}{c}{\textbf{{\begin{tabular}[c]{@{}c@{}}(b) Accuracies for \\ accumulated Rated sets\end{tabular}}}} \\ \cline{1-7} \cline{9-14} 
\multicolumn{1}{|l|}{\begin{tabular}[c|]{@{}l@{}}Trust Factor/\\ Tr. Set\end{tabular}} & 0.5              & 0.6              & 0.7              & 0.8              & 0.9             & 1               &  & 0.5               & 0.6               & 0.7              & 0.8              & 0.9              & 1                \\ \cline{1-7} \cline{9-14} 
\multicolumn{1}{|l|}{GT+PGT(1-2)}                                                     & 50.1             & 48.9             & 50.1             & 49.1             & 48.3            & 48.4            &  & 50.3              & 49.3              & 49.9             & 51               & 48               & 48.4             \\ \cline{1-7} \cline{9-14} 
\multicolumn{1}{|l|}{GT+PGT(1-3)}                                                     & 49.6             & 50.3             & 50               & 50.4             & 49.2            & 50.5            &  & 50.7              & 49.4              & 49.5             & 49.8             & 49.7             & 48.5             \\ \cline{1-7} \cline{9-14} 
\multicolumn{1}{|l|}{GT+PGT(1-4)}                                                     & 51.5             & 50               & 50               & 48.5             & 49.7            & 48.3            &  & 51.5              & 50.6              & 50.3             & 48.1             & 49.4             & 48               \\ \cline{1-7} \cline{9-14} 
\multicolumn{1}{|l|}{GT+PGT(1-5)}                                                     & 50.3             & 50.9             & 50.1             & 49.8             & 49.6            & 48.5            &  & 50.3              & 50.9              & 50.1             & 49.8             & 49.6             & 48.5             \\   \cline{1-7} \cline{9-14} 
\multicolumn{1}{|l|}{Average IoU}                                                     & 50.2             & 49.8             & 49.6             & 49.6             & 49.5            & 49.2            &  & 50.6              & 50.3              & 50.3             & 49.9             & 49.4             & 48.6             \\   \cline{1-7} \cline{9-14} 
\end{tabular}
\end{table}

{\textbf{Visual Rating:}}
The first columns of Tables~\ref{tab:acc_sets}(a,b) present an interesting observation. When the trust factor is sufficiently low ($t_f=0.5$), the accuracy for each set in Visually Rated accumulation is in general higher than or equal to the accuracy in case of Sequential accumulation. Fig. 6 illustrates this trend. The reason for lower accuracy of sequentially sorted accumulated sets is again due to label ambiguity as discussed before. For example, when the first and second sequentially sorted sets are added to the GT, there is a high correlation among the $1^{st}$, $2^{nd}$ and $3^{rd}$ images of the sequence and the effect of ambiguous labeling occurs (as discussed in Sec.~\ref{sec:amb_lab}). Of course such kind of correlated images are present in the visually sorted sets as well and would come up during accumulation.  But in the initial sets, for some GT images, there are no images of that sequence till PGT\_R3, PGT\_R4, PGT\_R5. This means there is no ambiguous labeling effect due to these images.
This brings us to a conclusion that visual rating helps when the PGT data is accumulated and the number of PGT images is multiple folds higher than GT images.  Evidently, accumulation does not explicitly help to enhance the accuracy over training with separated sets. The maximum accuracy achieved in both the cases is $51.5\%$. But clearly, this experiment suggests that such kind of accumulated training with sequences of images for semantic video  segmentation can help when there is usage of PGT data. Additionally, we conjecture that even better results can be obtained by adapting the trust factors to individual images based on label quality and other potential factors.


\vspace{-0.3cm}
\section{Conclusions}
\vspace{-0.3cm}
\label{sec:con}
In this work, we have explored the possibility of using \textit{pseudo ground truth} (PGT) to generate more training data for CNN. The main contribution is to systematically analyze three aspects of how the PGT has to be used to enhance the performance of a CNN-based semantic segmentation framework. From our experiments, we make the following conclusions: a) When the the number of PGT samples and GT samples is comparable, it is important to use diverse PGT data compared to GT images which also has good quality of labeling; b) When the number of PGT samples is multifold compared with GT, the trust on the PGT samples should be sufficiently low; c) Accumulation of PGT data does not explicitly help in improving the performance of semantic segmentation by a considerable amount. But it is important to note that in cases such as video processing, sequential labeled data has to be presented to the CNN and our experiments show that PGT can be used in such cases with a sufficiently low trust factor and it does not worsen the performance. To this end, we recommend that diverse high quality PGT should be used when one has to improve the performance of semantic segmentation. In case PGT is being used for semantic video segmentation, we recommend that the trust on PGT is kept to a low value. Additionally, there are many exciting avenues for future research. One direction is to improve the PGT data generation itself. As the experiments have shown, we believe that even better results can be achieved when the so-called trust factors are individually adapted to the data in case of PGT accumulation. For instance, each frame, or even each pixel, receives a different trust factor, potentially also conditioned on the image content and other information. Another direction of research is to compare and complement our approach for data augmentation to other common strategies for data augmentation.

\bibliographystyle{splncs03}
\bibliography{egbib}

\begin{thebibliography}{10}
\providecommand{\url}[1]{\texttt{#1}}
\providecommand{\urlprefix}{URL }

\bibitem{agrawal_nn_2014}
Agrawal, P., Girshick, R.B., Malik, J.: Analyzing the performance of multilayer
  neural networks for object recognition. In: ECCV. pp. 329--344 (2014)

\bibitem{BadrinarayananBC13}
Badrinarayanan, V., Budvytis, I., Cipolla, R.: Semi-supervised video
  segmentation using tree structured graphical models. PAMI  35(11),
  2751--2764 (2013)

\bibitem{BadrinarayananIJCV13}
Badrinarayanan, V., Budvytis, I., Cipolla, R.: Mixture of trees probabilistic
  graphical model for video segmentation. IJCV  110(1),  14--29 (2014)

\bibitem{BadrinarayananGC10}
Badrinarayanan, V., Galasso, F., Cipolla, R.: Label propagation in video
  sequences. In: CVPR (2010)

\bibitem{Brostow2008}
Brostow, G.J., Shotton, J., Fauqueur, J., Cipolla, R.: Segmentation and
  recognition using structure from motion point clouds. In: ECCV. pp. 44--57
  (2008)

\bibitem{corso_label_prop}
Chen, A.Y.C., Corso, J.J.: Propagating multi-class pixel labels throughout
  video frames. In: WNYIPW (2010)

\bibitem{Cordts2016Cityscapes}
Cordts, M., Omran, M., Ramos, S., Rehfeld, T., Enzweiler, M., Benenson, R.,
  Franke, U., Roth, S., Schiele, B.: The cityscapes dataset for semantic urban
  scene understanding. In: CVPR (2016)

\bibitem{EigenPF14}
Eigen, D., Puhrsch, C., Fergus, R.: Depth map prediction from a single image
  using a multi-scale deep network. In: NIPS. pp. 2366--2374 (2014)

\bibitem{Everingham:voc}
Everingham, M., Gool, L., Williams, C.K., Winn, J., Zisserman, A.: The pascal
  visual object classes (voc) challenge. IJCV  88(2),  303--338 (2010)

\bibitem{FragkiadakiCVPR12}
Fragkiadaki, K., Zhang, G., Shi, J.: Video segmentation by tracing
  discontinuities in a trajectory embedding. In: CVPR (2012)

\bibitem{Giordano2015}
Giordano, D., Murabito, F., Palazzo, S., Spampinato, C.: Superpixel-based video
  object segmentation using perceptual organization and location prior. In:
  CVPR (2015)

\bibitem{girshick2014rcnn}
Girshick, R., Donahue, J., Darrell, T., Malik, J.: Rich feature hierarchies for
  accurate object detection and semantic segmentation. In: CVPR (2014)

\bibitem{Godec2011}
Godec, M., Roth, P.M., Bischof, H.: Hough-based tracking of non-rigid objects.
  In: ICCV (2011)

\bibitem{gupta_ECCV14}
Gupta, S., Girshick, R., Arbelaez, P., Malik, J.: Learning rich features from
  {RGB-D} images for object detection and segmentation. In: ECCV. pp. 345--360
  (2014)

\bibitem{HartmannGHTKMVERS12}
Hartmann, G., Grundmann, M., Hoffman, J., Tsai, D., Kwatra, V., Madani, O.,
  Vijayanarasimhan, S., Essa, I.A., Rehg, J.M., Sukthankar, R.: Weakly
  supervised learning of object segmentations from web-scale video. In: ECCV
  (2012)

\bibitem{jain2014supervoxel}
Jain, S.D., Grauman, K.: Supervoxel-consistent foreground propagation in video.
  In: ECCV, pp. 656--671 (2014)

\bibitem{NIPS2015_5849}
Johnson, R., Zhang, T.: Semi-supervised convolutional neural networks for text
  categorization via region embedding. In: NIPS (2015)

\bibitem{KrahenbuhlK11}
Kr{\"{a}}henb{\"{u}}hl, P., Koltun, V.: Efficient inference in fully connected
  crfs with gaussian edge potentials. In: NIPS. pp. 109--117 (2011)

\bibitem{krizhevsky_cnn_2012}
Krizhevsky, A., Sutskever, I., Hinton, G.E.: Imagenet classification with deep
  convolutional neural networks. In: NIPS. pp. 1097--1105 (2012)

\bibitem{KunduLDLR14}
Kundu, A., Li, Y., Dellaert, F., Li, F., Rehg, J.M.: Joint semantic
  segmentation and 3d reconstruction from monocular video. In: ECCV (2014)

\bibitem{Lee2011}
Lee, Y.J., Kim, J., Grauman, K.: Key-segments for video object segmentation.
  In: ICCV (2011)

\bibitem{Li2013}
Li, F., Kim, T., Humayun, A., Tsai, D., Rehg, J.M.: Video segmentation by
  tracking many figure-ground segments. In: ICCV (2013)

\bibitem{Liu2015}
Liu, B., He, X.: Multiclass semantic video segmentation with object-level
  active inference. In: CVPR (2015)

\bibitem{long_shelhamer_fcn_2015}
Long, J., Shelhamer, E., Darrell, T.: Fully convolutional networks for semantic
  segmentation. In: CVPR (2015)

\bibitem{MobahiCW09}
Mobahi, H., Collobert, R., Weston, J.: Deep learning from temporal coherence in
  video. In: ICML (2009)

\bibitem{NagarajaSB15}
Nagaraja, N.S., Schmidt, F.R., Brox, T.: Video segmentation with just a few
  strokes. In: ICCV (2015)

\bibitem{Ochs2011}
Ochs, P., Brox, T.: Object segmentation in video: a hierarchical variational
  approach for turning point trajectories into dense regions. In: ICCV (2011)

\bibitem{oquab_cnn_2014}
Oquab, M., Bottou, L., Laptev, I., Sivic, J.: Learning and transferring
  mid-level image representations using convolutional neural networks. In:
  CVPR. pp. 1717--1724 (2014)

\bibitem{PapandreouCMY15}
Papandreou, G., Chen, L., Murphy, K.P., Yuille, A.L.: Weakly-and
  semi-supervised learning of a deep convolutional network for semantic image
  segmentation. In: ICCV (2015)

\bibitem{Papazoglou2013}
Papazoglou, A., Ferrari, V.: Fast object segmentation in unconstrained video.
  In: ICCV (2013)

\bibitem{PathakKD15}
Pathak, D., Kr{\"{a}}henb{\"{u}}hl, P., Darrell, T.: Constrained convolutional
  neural networks for weakly supervised segmentation. In: ICCV (2015)

\bibitem{pinheiro2014}
Pinheiro, P., Collobert, R.: Recurrent convolutional neural networks for scene
  labeling. In: ICML (2014)

\bibitem{Silberman:ECCV12}
Silberman, N., Hoiem, D., Kohli, P., Fergus, R.: Indoor segmentation and
  support inference from rgbd images. In: European Conference on Computer
  Vision (ECCV). pp. 746--760 (2012)

\bibitem{Tang2013}
Tang, K., Sukthankar, R., Yagnik, J., Fei-Fei, L.: Discriminative segment
  annotation in weakly labeled video. In: CVPR (2013)

\bibitem{toshev_pose_2014}
Toshev, A., Szegedy, C.: Deeppose: Human pose estimation via deep neural
  networks. In: CVPR. pp. 1653--1660 (2014)

\bibitem{Wu2015}
Wu, Z., Li, F., Sukthankar, R., Rehg, J.M.: Robust video segment proposals with
  painless occlusion handling. In: CVPR (2015)

\bibitem{XiaoXYHW15}
Xiao, T., Xia, T., Yang, Y., Huang, C., Wang, X.: Learning from massive noisy
  labeled data for image classification. In: CVPR (2015)

\bibitem{Yang2015}
Yang, Y., Sundaramoorthi, G., Soatto, S.: Self-occlusions and disocclusions in
  causal video object segmentation. In: ICCV (2015)

\bibitem{YaoTCBPLC15}
Yao, L., Torabi, A., Cho, K., Ballas, N., Pal, C.J., Larochelle, H., Courville,
  A.C.: Describing videos by exploiting temporal structure. In: ICCV (2015)

\bibitem{Zhang2013}
Zhang, D., Javed, O., Shah, M.: Video object segmentation through spatially
  accurate and temporally dense extraction of primary object regions. In: CVPR
  (2013)

\bibitem{zhang_rcnn_2014}
Zhang, N., Donahue, J., Girshick, R.B., Darrell, T.: Part-based r-cnns for
  fine-grained category detection. In: ECCV. pp. 834--849 (2014)

\end{thebibliography}
\end{document}